\pdfoutput=1 
 
\documentclass[10pt,twocolumn,letterpaper]{article}

\usepackage{cvpr}
\usepackage{times}
\usepackage{epsfig}
\usepackage{graphicx}
\usepackage{amsmath}
\usepackage{amssymb}
\usepackage{multirow}
\usepackage{bm}
\usepackage{mwe}
\usepackage{pifont}
\usepackage[abs]{overpic}

\newcommand{\fullname}{What-Where-When}%
\newcommand{\shortname}{W3}%
\newcommand{\nameFull}{\fullname{}}%
\newcommand{\keypoint}[1]{\vspace{0.1cm}\noindent\textbf{#1}\quad}
\newcommand{\cut}[1]{}

\usepackage{booktabs}

\usepackage[pagebackref=true,breaklinks=true,letterpaper=true,colorlinks,bookmarks=false]{hyperref}

\cvprfinalcopy 

\ifcvprfinal\fi
\begin{document}

\title{Egocentric Action Recognition by Video Attention and Temporal Context}


\author{
Juan-Manuel Perez-Rua$^{1}$ ~~ Antoine Toisoul$^{1}$ ~~ Brais Martinez$^{1}$ ~~ Victor Escorcia$^{1}$ \\
Li Zhang$^{1}$ ~~~ Xiatian Zhu$^{1}$ ~~~ Tao Xiang$^{1,2}$\\
{\small \textbf{[~j.perez-rua, a.toisoul, brais.a, v.castillo, li.zhang1, xiatian.zhu, tao.xiang~]@samsung.com}}\\
Samsung AI Centre, Cambridge\\
University of Surrey
}

\maketitle

\begin{abstract}
	We present the submission of Samsung AI Centre Cambridge to the CVPR2020 EPIC-Kitchens Action Recognition Challenge. 
	In this challenge, action recognition is posed as the problem of simultaneously predicting a single `verb' and `noun' class label given an input trimmed video clip. 
	That is, a `verb' and a `noun' together define a compositional `action' class.
	The challenging aspects of this real-life action recognition task include small fast moving objects, complex hand-object interactions, and occlusions.
	At the core of our submission
	is a recently-proposed spatial-temporal video attention model,
	called `W3' (`What-Where-When') attention~\cite{perez2020knowing}.
	We further introduce a simple yet effective contextual learning mechanism to model `action' class scores directly
	from long-term temporal behaviour 
	based on the `verb' and `noun' prediction scores. 
	Our solution achieves strong performance on the challenge metrics without using object-specific reasoning nor extra training data.
	In particular, our best solution with multimodal ensemble achieves the 2$^{nd}$ best position for `verb', and 3$^{rd}$ best for `noun' and `action' on the Seen Kitchens test set. 
\end{abstract}

\section{Introduction}

\emph{EPIC-Kitchens} is a large scale egocentric video benchmark for daily kitchen-centric activity understanding \cite{damen2018scaling}.
In this benchmark, 
the action classes are defined by combining 
verb and noun classes. 

By combining all the 352 nouns and 125 verbs, 
the number of all possible action classes will reach as large as 44,000. 
This dataset presents a long tail distribution
as often occurred in natural scenarios.
Besides, human-object interaction actions
might be very ambiguous. 
For example, in a single video clip, a person might be washing a dish whilst interacting with a sponge, faucet and/or sink concurrently, and sometimes 
the in-interaction active object might be completely occluded. 
These factors all render action recognition on this dataset extremely challenging.
%
%
Whilst significant progress has been made since the inception of this challenge~\cite{damen2018scaling,price2019evaluation},
it is rather clear from the performance of all previous winner solutions
that fine-grained action recognition is still far from being solved.

In this attempt, we present a novel egocentric action recognition solution
based on video attention learning
and temporal contextual learning jointly.
By focusing on the action class related regions in highly redundant video data over space and time, 
the model inference is made more robust against
noisy and distracting observations.
To this end, we exploit a recently-proposed \nameFull{} (W3) video attention model \cite{perez2020knowing}.
Temporal context provides additional useful information beyond individual video clips,
as there are inherent interdependent relationships 
of human actions in performing daily life activities.
For instance, it is more likely that a person is grabbing a cup if previously he/she was
opening a cupboard, than for example, if the person had just opened a washing machine. 
A Temporal Context Network (CtxtNet) is introduced to enhance model inference
by considering temporally adjacent actions
happening in a time window.

To make a stronger solution, we adopt multi-modal fusion, as in \cite{kazakos2019epic}, 
combining RGB (static appearance),
optical flow (motion cue), and audio information together.

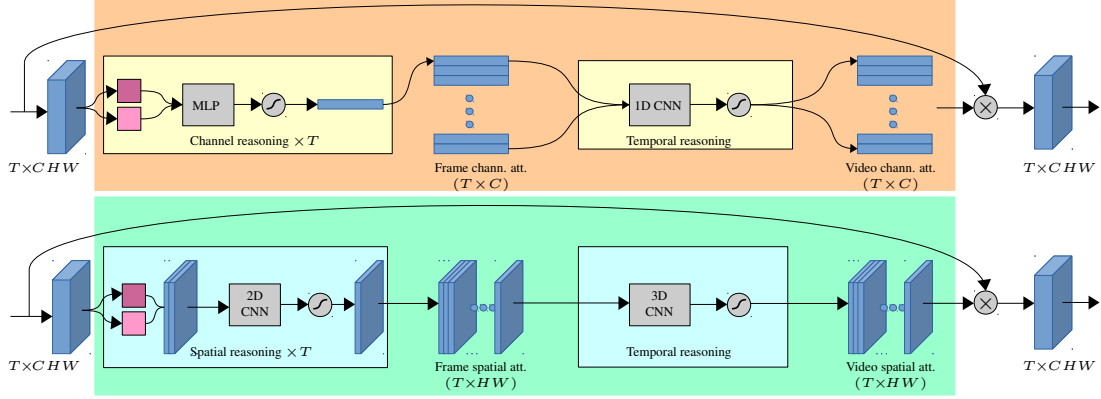
\begin{figure*}[ht]
	\centering
	\begin{overpic}[width=0.85\textwidth]{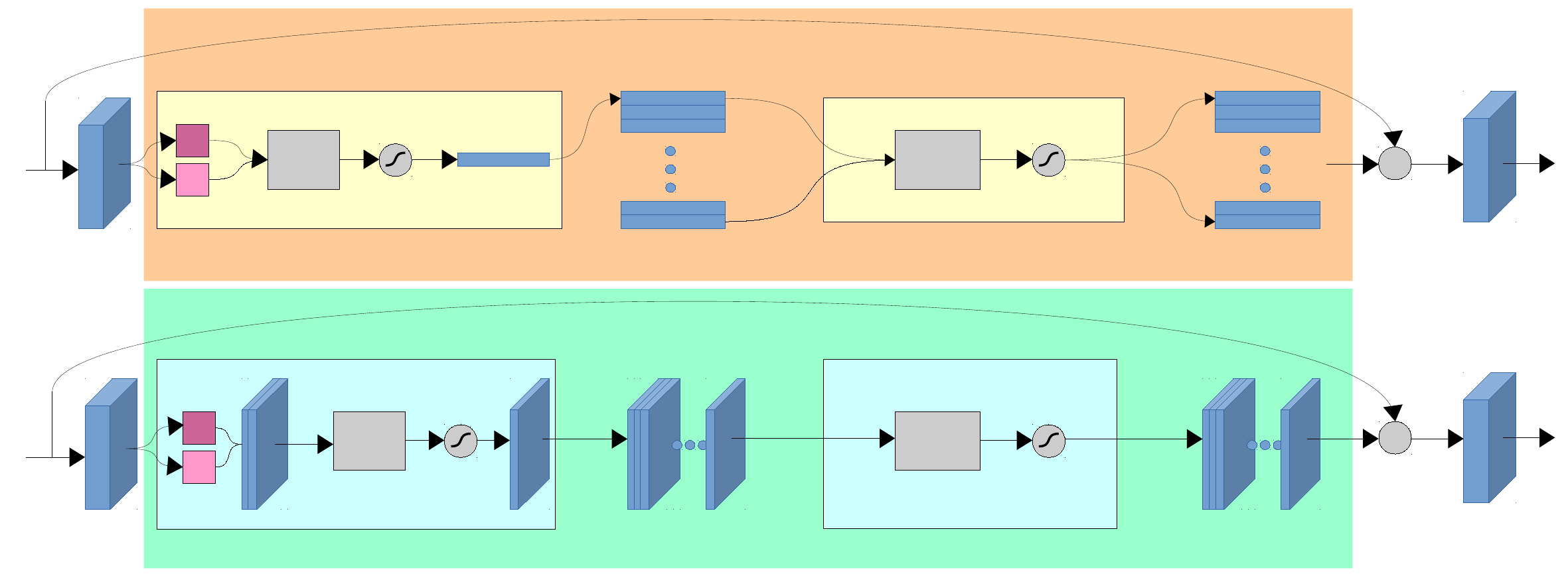}
		
		\put (6.0, 90) {\tiny{$T\hspace{-0.75mm}\times \hspace{-0.9mm} CHW$}}
		\put (6.0, 15) {\tiny{$T\hspace{-0.75mm}\times \hspace{-0.9mm} CHW$}}
		
		\put (61.0,  112.5) {\parbox{40pt}{\centering \tiny{MLP}}}
		\put (242.0, 112.5) {\parbox{20pt}{\centering \tiny{1D CNN}}}
		
		\put (75.0, 100) {\tiny{Channel reasoning $\times T$}}
		\put (240.0, 100) {\tiny{Temporal reasoning}}    
		
		\put (75, 20) {\tiny{Spatial reasoning $\times T$}}
		\put (240.0, 20) {\tiny{Temporal reasoning}}
		
		\put (94.0,  37.5) {\parbox{11pt}{\centering \tiny{2D CNN}}}
		\put (247.0,  37.5) {\parbox{11pt}{\centering \tiny{3D CNN}}}    
		
		\put (160.0,  85.5) {\parbox{50pt}{\centering \tiny{Frame chann. att. $(T\hspace{-0.5mm}\times\hspace{-0.5mm}C)$}}}
		\put (315.0,  85.5) {\parbox{50pt}{\centering \tiny{Video chann. att. $(T\hspace{-0.5mm}\times\hspace{-0.5mm} C)$}}}
		
		\put (315.0,  10.0) {\parbox{50pt}{\centering \tiny{Video spatial att. $(T\hspace{-0.75mm}\times\hspace{-0.9mm}HW)$}}}
		\put (160.0,  10.0) {\parbox{50pt}{\centering \tiny{Frame spatial att. $(T\hspace{-0.75mm}\times\hspace{-0.9mm}HW)$}}}
		
		\put (372.7,112.0) {\footnotesize{$\times$}}
		\put (372.7,38.0) {\footnotesize{$\times$}}    
		
		\put (390.0, 90.0) {\tiny{$T\hspace{-0.75mm}\times \hspace{-0.75mm} CHW$}}
		\put (390.0, 15.0) {\tiny{$T\hspace{-0.75mm}\times \hspace{-0.75mm} CHW$}}    
		
	\end{overpic}
	\caption{Schematic illustration of the \shortname{} attention module.
		{\bf Top:} The channel-temporal attention sub-module (orange box). 
		{\bf Bottom:} The spatial-temporal attention sub-module (green box). 
		The symbol $\bigotimes$ denotes point-wise multiplication between the attention maps and input features.
	}
	\label{fig:W3_detail}
\end{figure*}

\section{Methodology}

In this section, we present the solution 
of our submission to the \emph{EPIC-Kitchens Action Recognition challenge}.
We first introduce the W3 attention model~\cite{perez2020knowing} in Section~\ref{sec:w3}, and then describe our proposed temporal action context model (CtxtNet) in Section~\ref{sec:tcn}.

\subsection{What-Where-When Attention}

\label{sec:w3}
Figure \ref{fig:W3_detail} gives the schematic illustration of the \shortname{} attention module.
Significantly, \shortname{} can be plugged into any existing video action recognition network, \eg TSM~\cite{lin2019tsm}, for end-to-end learning.
Specifically, \shortname{} accepts a single feature map ${\bf F}$ as input, which can be derived from any CNN layer, 
and generates an attention map ${\bf M}$ with same dimension to ${\bf F}$, \ie, ${\bf F},{\bf M} \in \mathbb{R}^{T\times C\times H \times W}$,
where $T, C, H, W$ denote the number of video clip frame, 
number of feature channel, 
height and width of the frame-level feature map respectively. 
Attention mask ${\bf M}$ is then used to produce a refined feature map ${\bf F}'$ in a way that only action class-discriminative cues are allowed to flow forward, whilst irrelevant ones are suppressed.
The attention and refined feature learning process is expressed as:

\begin{equation}
\label{eq:w3}
{\bf F}' = {\bf F} \otimes {\bf M}, \;\;\;\; {\bf M} = f({\bf F}),
\end{equation}

where $\otimes$ is the Hadamard product, and $f(.)$ is the W3 attention function. 

To facilitate effective and efficient attention learning, 
\shortname{} adopts an attention factorization scheme
by splitting the 4D attention tensor ${\bf M}$ into
a channel-temporal attention mask ${\bf M}^c \in \mathbb{R}^{T\times C}$
and 
a spatial-temporal attention mask ${\bf M}^s \in \mathbb{R}^{T\times H \times W}$.
This strategy reduces the complexity of the learning problem as the size of the 
attention masks are reduced from $TCHW$ to $T(C+HW)$.
In principle, the feature attending scheme in Eq~\ref{eq:w3} is thus reformulated into a two-step sequential process:

\begin{equation}
{\bf F}^c = {\bf M}^c \otimes {\bf F}, \; {\bf M}^c = f^c({\bf F}); \;\;\;\;
\end{equation}

\begin{equation}
{\bf F}^s = {\bf M}^s \otimes {\bf F}^c, \; {\bf M}^s = f^s({\bf F}^c);
\end{equation}

where $f^c(.)$ and $f^s(.)$ denote the channel-temporal and spatial-temporal attention function respectively.

\keypoint{Channel-temporal attention}
\label{sec:channel_att}
The channel-temporal attention focuses on the `what-when' facets of video attention. 
Specifically it measures the importance of a particular  object-motion pattern evolving temporally across a video sequence in a specific way. 
For this, we squeeze the spatial dimensions ($H\times W$) of each frame-level 3D feature map 
to yield a compact channel descriptor ${\bf d}_{\text{chnl}} \in \mathbb{R}^{T\times C}$
as in \cite{hu2018squeeze}.
Moreover, we use both max and mean pooling operations as in \cite{woo2018cbam}, and denote the two channel descriptors as ${\bf d}_{\text{avg-c}}$ and ${\bf d}_{\text{max-c}} \in \mathbb{R}^{C\times 1 \times 1}$ (indicated by the purple boxes in the top of Fig.~\ref{fig:W3_detail}).

To extract the inter-channel relationships for a given frame,
we then forward ${\bf d}_{\text{avg-c}}$
and ${\bf d}_{\text{max-c}}$ into a MLP $\theta_{\text{c-frm}}$.
The above {\em fWhatrame-level channel-temporal attention} can be expressed as:

\begin{equation}
{\bf M}^{\text{c-frm}} = 
\sigma\Big(f_{\theta_{\text{c-frm}}}({\bf d}_{\text{avg-c}}) \oplus f_{\theta_{\text{c-frm}}}({\bf d}_{\text{max-c}})\Big)
\in \mathbb{R}^{C\times 1 \times 1},
\end{equation}

where $f_{\theta_{\text{c-frm}}}(.)$ outputs channel frame attention
and $\sigma(.)$ is the sigmoid function.

In EPIC-Kitchens, it is critical to model the temporal dynamics of active objects in interaction with the human subject.
To capture this information, a small channel temporal attention network ${\theta_{\text{c-vid}}}$ is introduced, composed of 
a CNN network with two layers of 1D convolutions, to reason about the 
temporally evolving characteristics of each channel dimension (Fig.~\ref{fig:W3_detail} top-right).
This results in our channel-temporal attention mask ${\bf M}^c$, computed as:

\begin{equation}
{\bf M}^{\text{c}} = 
\sigma\Big(f_{\theta_{\text{c-vid}}}(\{{\bf M}^{\text{c-frm}}_i\}_{i=1}^T)\Big).
\end{equation}

Concretely, this models the per-channel temporal relationships of successive frames  in a local window specified 
by the kernel size $K_\text{c-vid}$, and composed by two layers.

\keypoint{Spatial-temporal attention}
\label{sec:spatial_att}
In contrast to the channel-temporal attention that attends to dynamic object feature patterns evolving temporally in certain ways, this sub-module attempts to localize them over time.
Similarly to the previous module, we apply average-pooling and max-pooling along the channel axis to obtain two compact 2D spatial feature maps
for each video frame, denoted as ${\bf d}_{\text{avg-s}}$
and ${\bf d}_{\text{max-s}} \in \mathbb{R}^{1\times H\times W}$.
We then concatenate the two maps and deploy a spatial attention network $\theta_{\text{s-frm}}$ with one 2D convolutional layer 
for each individual frame to output the frame-level spatial attention ${\bf M}^{\text{s-frm}}$ (see Fig.~\ref{fig:W3_detail} bottom-left).
To incorporate the temporal dynamics to model how spatial attention evolves over time,
we further perform temporal reasoning on $\{ {\bf M}_i^{\text{s-frm}}\}_{i=1}^T \in \mathbb{R}^{T \times H \times W}$ using a lightweight 3D CNN $\theta_{\text{s-vid}}$.
We adopt a kernel size of $3\times 3\times 3$~(Fig.~\ref{fig:W3_detail} bottom-right).
The {\em frame-level and video-level spatial attention} are, then:

\begin{equation}
{\bf M}^{\text{s-frm}} = 
\sigma\Big(f_{\theta_{\text{s-frm}}}([{\bf d}_{\text{avg-s}},{\bf d}_{\text{max-s}}])\Big)
\in \mathbb{R}^{1\times H \times W},
\end{equation}

\begin{figure}[t]
	\centering
	\includegraphics[width=1.0\columnwidth]{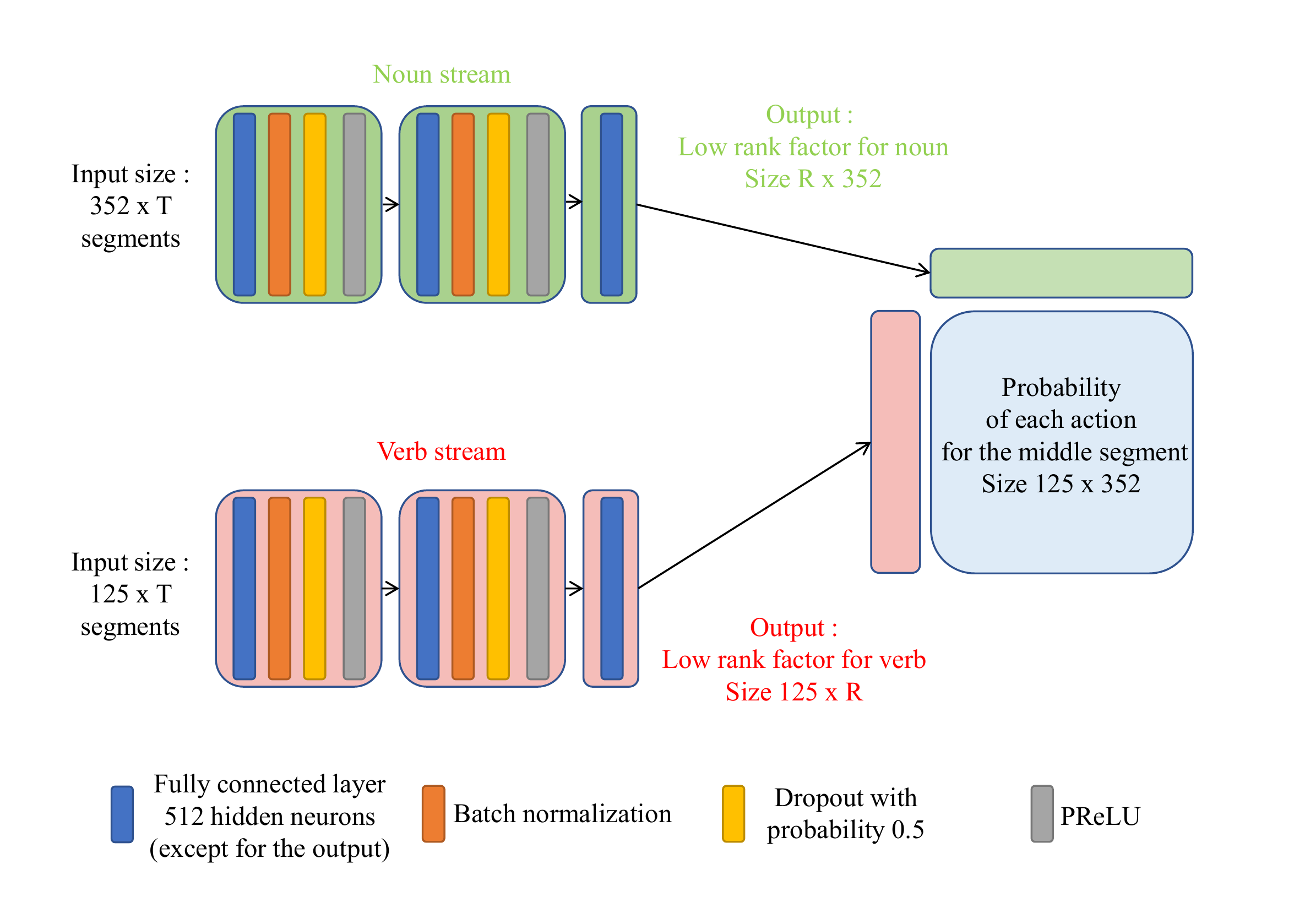}
	
	\caption{{\bf Overview of the proposed Temporal Context Network (CtxtNet).} Noun and verb predictions
		are incorporated through a low-rank factorisation scheme to produce the final
		action scores.
	}
	\label{fig:ctxt}
\end{figure}

\begin{equation}
{\bf M}^{\text{s}} = 
\sigma\Big(f_{\theta_{\text{s-vid}}}(\{{\bf M}^{\text{s-frm}}_i\}_{i=1}^T)\Big) \in \mathbb{R}^{T \times H \times W}.
\end{equation}

\subsection{Temporal Context Network}
\label{sec:tcn}

The objective of 
contextual learning
is to provide a per-clip action prediction
by taking into account the surrounding actions (their verb and noun component),
i.e., temporal context.
This brings in additional information source on top of the
observation of isolated short video clips.

For instance, the action ``open cupboard'' is more likely to be followed by ``close cupboard'', as compared with ``cut onions''.

A straightforward method is to learn a non-linear mapping from the combination of verb and noun predictions to the action class label space.
However, this is computationally not tractable due to the huge action spaces with 44000 class labels, which also runs a high risk of overfitting.

To alleviate these issues, we propose to learn a low rank factorisation of the action matrix to more efficiently encode context information by designing a Temporal Context Network (CtxtNet).
An overview of CtxtNet is shown in Fig.~\ref{fig:ctxt}.

In particular, CtxtNet is made of two parallel 3-layer MLP streams, one for noun and one for verb. 
Each stream generates a low rank matrix (of size $125 \times R$ for verbs and $R \times 352$ for noun)
whose multiplication yields 
the probability of each action, i.e., the action matrix of size $125 \times 352$. 
The hyperparameter $R$ controls the rank of the factorisation so as allowing to choose the trade-off between complexity of the reconstruction (the number of parameters) and the model capacity.
To encode the spatio-temporal context, the two streams act on a temporal context of $T$ frames.
In practice, we found that rank $R=16$ and a time window $T=5$ leads to the best results on a held-out validation set. 

\begin{table}[th]
	\centering
	\setlength{\tabcolsep}{0.1cm} 
	\resizebox{\columnwidth}{!}{
		\begin{tabular}{c|cc|cc|cc|cc|cc|cc}
			\toprule
			& \multicolumn{4}{c|}{ {\bf Verb} } & \multicolumn{4}{c|}{ {\bf Noun} }  
			& \multicolumn{4}{c}{ {\bf Action} }  \\
			
			& \multicolumn{2}{c|}{Top-1} & \multicolumn{2}{c|}{Top-5}  
			& \multicolumn{2}{c|}{Top-1} & \multicolumn{2}{c|}{Top-5}
			& \multicolumn{2}{c|}{Top-1} & \multicolumn{2}{c}{Top-5} \\
			
			{\bf Model} & S1 & S2 & S1 & S2 & S1 & S2 & S1 & S2 & S1 & S2 & S1 & S2  \\
			\midrule
			TSM \cite{lin2019tsm} 
			& 57.9 & 43.5 & 87.1 & 73.9 & 40.8 & 23.3 & 66.1 & 46.0 & 28.2 & 15.0 & 49.1 & 28.1 \\
			TSM+NL \cite{wang2018non} & 60.1 & 49.0 & 87.3 & 77.5 & 42.8 &\bf 27.7 & 66.4 &\bf 51.2 & 30.8 & 18.0 & 50.0 & 33.0 \\		
			TSM+W2
			& 63.4 & 50.0 & \bf 88.8 & 77.0 & 44.3 & 26.7 & 68.6 & 50.0 & 33.2 & 17.9 &\bf 54.6 & 32.7 \\
			\hline
			TSM+{\bf \shortname{}}
			&\bf  64.7 &\bf  51.4 &\bf 88.8 &\bf 78.5 & \bf 44.7 & 27.0 & \bf 69.0 & 50.3 &\bf 34.2 & \bf 18.7 & \bf 54.6 & \bf 33.7
			\\
			\bottomrule
		\end{tabular}
	}
	\caption{\footnotesize
		{\bf TSM with different attention modules}. 
		Setting: 8 frames per video, only RGB frames.
		Backbone: ResNet-50~\cite{he2016deep}.
		S1: Seen Kitchens;
		S2: Unseen Kitchens.
		W2: W3 without temporal component.
		Experiments run with 10-crops and 2 clips per-video.
	}
	\label{tab:epicatt}
\end{table}

\begin{table}[th]
	\footnotesize
	\centering
	\setlength{\tabcolsep}{0.1cm} 
	
	\begin{tabular}{c|c|c|c}
		\toprule
		\bf Model & \bf Verb & \bf Noun & \bf Action\\
		\midrule
		TSM ResNet-50
		& 58.88 & 42.74 & 30.40 \\
		TSM ResNet-101
		& 62.14 & 45.16 & 34.28 \\
		TSM ResNet-152
		& 63.39 & 45.70 & 34.78 \\
		\midrule
		TSM ResNet-152 + {\bf W3}
		& 62.64 & 46.66 & 36.86
		\\
		\bottomrule
	\end{tabular}
	
	\caption{\footnotesize
		{\bf TSM using different ResNet backbones on the validation set}. 
		Setting: 8 frames per video, only RGB frames used.
	}
	\label{tab:epicbackbone}
\end{table}

\begin{table}[ht]
	\footnotesize
	\centering
	\setlength{\tabcolsep}{0.1cm} 
	\begin{tabular}{c|cc|cc|cc}
		\toprule
		& \multicolumn{2}{c|}{ {\bf Verb} } & \multicolumn{2}{c|}{ {\bf Noun} }  
		& \multicolumn{2}{c}{ {\bf Action} }  \\
		\midrule
		& \multicolumn{2}{c|}{Top-1 Acc.}
		& \multicolumn{2}{c|}{Top-1 Acc.} 
		& \multicolumn{2}{c}{Top-1 Acc.} \\
		& S1 & S2 & S1 & S2 & S1 & S2  \\
		\midrule
		Action Prior~\cite{price2019evaluation} &
		69.18  & 57.76  &
		49.58  & 33.59  &
		38.12  & 23.72  \\
		CtxtNet (Ours) &
		69.18  & 57.76  &
		49.58  &  33.59  &
		\bf 39.30  & 23.38  \\		
		\bottomrule
	\end{tabular}
	\caption{\footnotesize
		{\bf Effect of CtxtNet}. 
		S1: Seen Kitchens;
		S2: Unseen Kitchens.
		Results obtained in the EPIC-Kitchens test server.
	}
	\label{tab:epicctxt}
\end{table}

\section{Experiments}

\begin{table*}[!t]
	\footnotesize
	\centering
	\setlength{\tabcolsep}{0.1cm} 
	\begin{tabular}{c|cc|cc|cc|cc|cc|cc}
		\toprule
		& \multicolumn{4}{c|}{ {\bf Verb} } & \multicolumn{4}{c|}{ {\bf Noun} }  
		& \multicolumn{4}{c}{ {\bf Action} }  \\
		\midrule
		& \multicolumn{2}{c|}{Top-1} & \multicolumn{2}{c|}{Top-5}  
		& \multicolumn{2}{c|}{Top-1} & \multicolumn{2}{c|}{Top-5}
		& \multicolumn{2}{c|}{Top-1} & \multicolumn{2}{c}{Top-5} \\
		
		& S1 & S2 & S1 & S2 & S1 & S2 & S1 & S2 & S1 & S2 & S1 & S2  \\
		\midrule
		All modalities &
		69.43 (2) & 57.60 (4) &
		91.23 (2) & 81.84 (4) &
		
		49.71 (3) & 34.69 (4) &
		73.18 (3) & 61.25 (3) &
		
		40.00 (3) & 24.62 (6) &
		60.53 (3) & 41.38 (6) \\
		\bottomrule
	\end{tabular}
	\caption{\footnotesize
		{\bf Final scores on the testing server}. 
		Setting: 8 frames per video.
		Modalities: RGB-W3, RGB, Flow, Audio.
		Backbone: ResNet-50~\cite{he2016deep}.
		S1: Seen Kitchens;
		S2: Unseen Kitchens.
		Results obtained in the EPIC-Kitchens test server with 1 crop and 2 clips per-video.
		(X): Position in the 2020 ranking.
	}
	\label{tab:epicfinal}
\end{table*}

\keypoint{Setup}
In the video classification track of EPIC-Kitchens,
there are three classification tasks involved:
noun classification, verb classification,
and their combination.
Two different held-out testing sets
are considered: 

Seen Kitchens Testing Set (S1), and Unseen Kitchens Testing Set (S2).

\keypoint{Validation set} 
To allow for apples-to-apples comparison to other methods, we used the same validation set as~\cite{kazakos2019epic}.

\keypoint{Experimental details}
We used the recent Temporal Shift Module (TSM)~\cite{lin2019tsm} as the baseline video recognition model in all the experiments. We trained our models for 50 epochs with SGD, at a learning rate of 0.02.
The models were initialized by ImageNet pre-training.
Unless otherwise mentioned, our default backbone network is a ResNet-152. W3 models were trained with mature feature regularisation (MFR) as described in~\cite{perez2020knowing}.
The CtxtNet was trained with Adam in a second stage. Firstly, we employed our multi-modal ensemble to compute the verb and noun logits of each video segment. The CtxtNet then maps those verb and noun logits to an action probability matrix. 
For both branches of CtxtNet, the MLP is a stack of three linear layers. Each of them was formed by a linear projection, batch norm, PReLU and Dropout.
For all the experiments, unless otherwise mentioned, we sampled two clips
and a single central crop per video.
Finally, for our last submission, we assembled two models per modality, except for audio, for which we only used a single model.

\subsection{Attention Model Comparison} 

We compared our W3 attention with existing competitive
alternatives.
For fair comparison experiments, all attention methods use the same ResNet-50 based TSM~\cite{lin2019tsm} as the underlying video model.

Table~\ref{tab:epicatt} shows that our W3 attention 
module is the strongest amongst several competitors.

\subsection{Backbone Network Evaluation} 

We tested our method with different backbone networks.
Table~\ref{tab:epicbackbone} shows that
ResNet-152~\cite{he2016deep} is slightly better than
ResNet-101, and almost five points better than ResNet-50. 
Importantly, it is shown that W3 further brings extra model performance improvement on top of the strongest backbone
on noun and action classification.
However, we observed that the performance of verb is not benefited from W3. 
This can be exploited by using both type of models for the RGB modality.

\subsection{Temporal Context Network}

Table~\ref{tab:epicctxt} shows that
our CtxtNet module produces better scores than the action prior method introduced by~\cite{price2019evaluation}.
CtxtNet brings a large gain on the seen kitchen setting at a small cost on the unseen kitchen setting.
Note, noun and verb accuracy scores are unaffected
since this does not change their predictions. 

\subsection{Multi-Modalities}

Table~\ref{tab:epicfinal} reports the final results of our method
using three modalities: 
RGB, optical flow, and audio.
This was made by a logit-level ensemble of
regular RGB model (ResNet-152),  
W3-attended RGB model (ResNet-152), 
optical flow model (ResNet-152), 
and audio model (ResNet-34).
For audio, we used spectrograms with the same format of~\cite{kazakos2019epic}.

\section{Conclusion}

In this report,
we summarised the model designs and implementation details of our solution
for video action classification.
With the help of the proposed W3 video attention 
and temporal context learning,
we achieved top-3 video action classification performance on the leaderboard.

{\footnotesize
\bibliographystyle{ieee_fullname}
\bibliography{egbib}
}

\end{document}